\documentclass[accepted]{uai2021} 

\usepackage[american]{babel}

\usepackage{natbib} 
\usepackage{mathtools} 
\usepackage{amssymb}
\usepackage{booktabs} 
\usepackage{tikz} 
\usepackage{cleveref}
\usepackage{subcaption}
\usepackage{hyperref}
\usepackage{relsize}
\usepackage{enumitem}
\hypersetup{colorlinks=True, allcolors=blue}

\usepackage{amsthm}
\newtheorem{definition}{Definition}
\usepackage{bbm}

\usepackage{balance}

\newcommand{\given}{\,|\,}
\newcommand{\compcond}[1]{\big(#1\given-\big)}

\newcommand{\teq}{\!=\!}
\newcommand{\tp}{\!+\!}
\newcommand{\tm}{\!-\!}
\newcommand{\tsim}{\!\sim\!}

\newcommand{\eps}[1]{\epsilon_0^{\mathsmaller{(#1)}}}
\newcommand{\lamij}[1]{\lambda_{ij}^{\mathsmaller{(#1)}}}
\newcommand{\yij}[1]{y_{ij}^{\mathsmaller{(#1)}}}
\newcommand{\thetaik}[1]{\theta_{ik}^{\mathsmaller{(#1)}}}
\newcommand{\gammaij}[1]{\gamma_{ij}^{\mathsmaller{(#1)}}}
\newcommand{\alphaij}[1]{\alpha_{ij}^{\mathsmaller{(#1)}}}

\newcommand{\Pois}[1]{\textrm{Pois}\left( #1 \right)}
\newcommand{\Bess}[1]{\textrm{Bess}\left( #1 \right)}

\newcommand{\Gam}[1]{\textrm{Gam}\left( #1 \right)}

\newcommand{\Multi}[1]{\textrm{Multi}\left( #1 \right)}

\newcommand{\DNCB}[1]{\textrm{DNCB}\left( #1 \right)}

\usepackage{xr-hyper}
\makeatletter
\newcommand*{\addFileDependency}[1]{
  \typeout{(#1)}
  \@addtofilelist{#1}
  \IfFileExists{#1}{}{\typeout{No file #1.}}
}
\makeatother

\newcommand*{\myexternaldocument}[1]{%
    \externaldocument{#1}%
    \addFileDependency{#1.tex}%
    \addFileDependency{#1.aux}%
}
\myexternaldocument{schein_713-supp}

\title{Doubly Non-Central Beta Matrix Factorization for DNA Methylation Data}

\author[1]{\href{mailto:Aaron Schein <aaron.schein@columbia.edu>?Subject=Your UAI 2021 paper}{Aaron Schein}{}} 
\author[2]{Anjali Nagulpally}
\author[3]{Hanna Wallach}
\author[2]{Patrick Flaherty}
\affil[1]{%
    Data Science Institute\\
    Columbia University
}
\affil[2]{%
    Department of Mathematics and Statistics\\
    University of Massachusetts Amherst
}

\affil[3]{%
    Microsoft\\
    New York City, NY
}

\begin{document}
\maketitle

\begin{abstract}
  We present a new non-negative matrix factorization model for $(0,1)$
  bounded-support data based on the doubly non-central beta (DNCB)
  distribution, a generalization of the beta distribution. The
  expressiveness of the DNCB distribution is particularly useful for
  modeling DNA methylation datasets, which are typically highly
  dispersed and multi-modal; however, the model structure is
  sufficiently general that it can be adapted to many other domains
  where latent representations of $(0,1)$ bounded-support data are of
  interest. Although the DNCB distribution lacks a closed-form
  conjugate prior, several augmentations let us derive an efficient
  posterior inference algorithm composed entirely of analytic updates.
  Our model improves out-of-sample predictive performance on both real
  and synthetic DNA methylation datasets over state-of-the-art methods
  in bioinformatics. In addition, our model yields meaningful latent
  representations that~accord with existing biological
  knowledge.\looseness=-1
\end{abstract}

\vspace{-0.55em}
\section{Introduction}
\label{sec:intro}
\vspace{-0.55em}

DNA methylation is a mechanism by which epigenetic changes to DNA can
modify the transcription of nearby genes.  These epigenetic changes
can activate oncogenes or inactivate tumor suppressors to drive the
onset of cancer and other
diseases~\citep{laird2010principles}. Discovering novel subtypes of
cancer that share underlying patterns of DNA methylation is of
interest to scientists, who seek to better understand the role of DNA
methylation in cancer development, and to clinicians, who seek to
refine existing cancer treatment strategies. To achieve this goal,
computational biologists routinely apply dimensionality reduction
methods to DNA methylation datasets in order to discover latent
representations that~are~both scientifically interesting and
clinically useful.\looseness=-1

A DNA methylation dataset typically consists of an $N \times M$
sample-by-gene matrix of bounded-support data $\mathcal{B} \in
(0,1)^{N \times M}$, where the number of samples $N$ is often far
exceeded by the number of genes $M$. A single element of this matrix
$\beta_{ij} \in (0,1)$ represents the degree of methylation~for
regions of the genome near gene $j$ in sample $i$. \looseness=-1

The most commonly used dimensionality reduction methods are principal
component analysis (PCA)~\citep{teschendorff2007elucidating} and
non-negative matrix factorization
(NMF)~\citep{zhuang2012comparison}. These methods are based on
Gaussian assumptions that are inappropriate for $(0,1)$ bounded-support data.
As a result, they fit DNA methylation datasets worse than methods that
are based on more appropriate~probabilistic
assumptions~\citep{ma2014comparisons}.\looseness=-1

The few existing non-Gaussian dimensionality reduction methods for DNA
methylation datasets almost all assume that the elements of a
sample-by-gene matrix are beta-distributed. Indeed, this assumption is
so standard in bioinformatics that the elements are typically referred
to as ``beta values''~\citep{kuan2010statistical}. Of these existing
methods, the most expressive is beta-gamma non-negative matrix
factorization (BG-NMF), a non-negative matrix factorization model with
a beta likelihood~\citep{ma2014comparisons}.\looseness=-1

Although the beta distribution is a natural choice for modeling data
with bounded support between 0 and 1, it is a challenging distribution
with which to build probabilistic models due its lack of a closed-form
conjugate prior~\citep{fink1997compendium}. In general, there are few
tractable and modular motifs for deriving posterior inference
algorithms for models that assume a beta likelihood. Models that do
not make overly simplistic assumptions tend to be accompanied by
posterior inference algorithms that are highly tailored to their
specific structures, making them difficult to modify or
extend. Moreover, these inference algorithms typically rely on
approximations that hamper precise quantification of uncertainty,
which is of particular interest in biomedical settings where datasets
are often small and properly calibrated decisions are
critical.\looseness=-1

\pagebreak

\begin{figure}[t!]
  \centering\textbf{}
  \includegraphics[width=\linewidth]{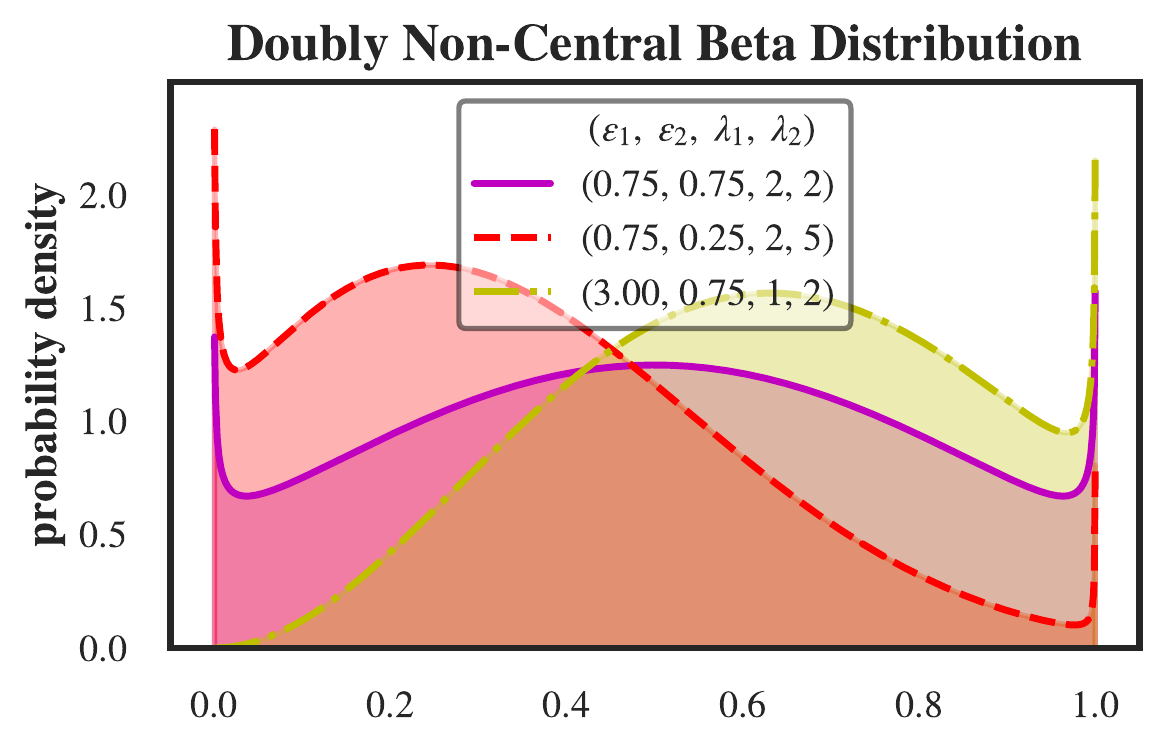}
\caption{\footnotesize The DNCB distribution can take multi-modal
    shapes when $\epsilon_1 < 1$ or $\epsilon_2 < 1$. This expressiveness
    is particularly useful when modeling DNA methylation datasets,
    which are typically highly dispersed and multi-modal. The DNCB
    distribution can~also~look like the standard beta
    (see \cref{fig:dncb_unimodal}~of the appendix).}\label{fig:dncb}
    \vspace{-1\baselineskip}
\end{figure}

In this paper, we therefore introduce a new non-negative matrix
factorization model for $(0,1)$ bounded-support data based on the
doubly non-central beta (DNCB)
distribution~\citep{ongaro2015some}. The DNCB distribution is a
four-parameter generalization of the beta distribution that can take a
more flexible set of shapes over the $(0,1)$ interval, including those
that are multi-modal (see \cref{fig:dncb}). Although our model was
developed specifically for DNA methylation datasets, the model
structure is sufficiently general that it can be adapted to many other
domains where latent representations of $(0,1)$ bounded-support data
are of interest.\looseness=-1

The property of the DNCB distribution that makes it particularly
useful for building probabilistic models is that it can be augmented
in terms of a pair of Poisson-distributed auxiliary variables. With
this augmentation, we can build tractable dimensionality reduction
methods for $(0,1)$ bounded-support data based on Poisson
factorization models, which are well studied and easy to build
on~\citep{titsias2007infinite,cemgil09bayesian,zhou2012lognormal,gopalan12scalable,paisley2014bayesian}.\looseness=-1

We develop an accompanying Gibbs sampler by appealing to
special relationships between the beta, gamma, and Poisson
distributions to obtain analytic updates that involve the Bessel
distribution~\citep{yuan2000bessel}. Our Gibbs sampler is
asymptotically guaranteed to sample from the exact posterior
distribution and it is general to any Poisson factorization model for
which analytic updates already exist.\looseness=-1

We compare our model's out-of-sample predictive performance to that of
state-of-the-art methods in bioinformatics. We find that our model
performs significantly better than NMF, which assumes a Gaussian
likelihood, and BG-NMF, which assumes a beta likelihood, for real DNA
methylation datasets. We also use a biologically motivated synthetic
data generator~\citep{de2020epiclomal} to create synthetic datasets
that enable us to study our model's suitability for
$(0,1)$~bounded-support data that may arise in other
domains.\looseness=-1

Finally, we explore our model's ability to discover meaningful latent
representations by applying our model to a microarray dataset composed
of samples from six different cancer types. We demonstrate that the
resulting representations accord with existing epigenetic knowledge
about the gene pathways that play major roles in the six different
cancer types.\looseness=-1

\paragraph{Contributions and roadmap.} To summarize, in this paper, we make three main contributions, outlined below:
\begin{enumerate}[leftmargin=*]
\item A new non-negative matrix factorization model for data with
  bounded support between 0 and 1 based on the doubly non-central beta
  (DNCB) distribution (\cref{sec:model}).\looseness=-1
\item An auxiliary variable scheme, involving several augmentations, that lets us develop an efficient Gibbs sampler
  composed entirely of analytic updates (\cref{sec:mcmc}).\looseness=-1
    \item A study of our model's out-of-sample predictive performance
      on real and synthetic DNA methylation datasets
      (\cref{sec:predictive}), and a case study demonstrating that the
      model also yields meaningful latent representations that accord with
      existing biological knowledge
      (\cref{sec:qualitative}).\looseness=-1
\end{enumerate}

\vspace{-0.55em}
\section{DNCB matrix factorization}
\label{sec:model}
\vspace{-0.55em}

Here, we present doubly non-central beta matrix factorization
(DNCB-MF), a new model that assumes each element $\beta_{ij} \in
(0,1)$ in a sample-by-gene matrix is drawn as follows:\looseness=-1
\begin{align}
\label{eq:marglikelihood}
    \beta_{ij} &\sim \DNCB{\eps{1}, \eps{2}, \lamij{1}, \lamij{2}},
\end{align}
where $\eps{1}$ and $\eps{2}$ are shared across all $i$ and $j$ and
$\lamij{1}$ and $\lamij{2}$ are linear functions of low-rank latent
factors---i.e.,\looseness=-1
\begin{align}
\label{eq:factorize}
    \lamij{1} = \sum_{k=1}^K \thetaik{1} \phi_{kj} &\;\;\;\;\textrm{and}\;\;\;\;
    \lamij{2} = \sum_{k=1}^K \thetaik{2} \phi_{kj}.
\end{align}
DNCB-MF is one instance of a class of models for $(0,1)$
bounded-support data that factorize the ``non-centrality'' parameters
of the DNCB distribution, as defined below.\looseness=-1
\begin{definition}
\label{def:dncb}
The doubly non-central beta (DNCB) distribution is continuous over the
support $\beta \in (0,1)$ and defined by ``shape'' parameters
$\epsilon_1, \epsilon_2 > 0$, ``non-centrality'' parameters
$\lambda_1, \lambda_2 \geq 0$, and the following probability density function:~\looseness=-1
  \begin{align}
      &\nonumber\textrm{DNCB}\,(\beta \,;\,\epsilon_1, \epsilon_2, \lambda_1, \lambda_2) \\
      &\nonumber\hspace{0.6em}= \text{Beta}\left(\beta;\, \epsilon_1, \epsilon_2 \right)\, e^{-\lambda_\bullet}\, \Psi_2\big[\epsilon_\bullet; \epsilon_1, \epsilon_2;\, \lambda_1\beta,\, \lambda_2 (1\!-\!\beta)\big],
  \end{align}
  where $\lambda_\bullet = \lambda_1+\lambda_2$, $\epsilon_\bullet =
  \epsilon_1+\epsilon_2$, and $\Psi_2[\cdot; \cdot, \cdot; \cdot,
    \cdot]$ denotes Humbert's confluent hypergeometric function~\citep{srivastava1985multiple,ongaro2015some}.\looseness=-1
\end{definition}

The key property of the DNCB distribution that makes it particularly
useful for building probabilistic models is that it can be augmented
in terms of a pair of Poisson-distributed~auxiliary variables, as
defined below.\looseness=-1

\begin{definition}
\label{def:prbeta}
A random variable $\beta \!\sim\! \textrm{DNCB}(\epsilon_1,
\epsilon_2, \lambda_1, \lambda_2)$ can be drawn from a standard beta
distribution conditioned on two Poisson-distributed auxiliary
variables as follows:\looseness=-1
\begin{align}
    y_1 \sim \Pois{\lambda_1} &\;\;\;\;\textrm{and}\;\;\;\;
    y_2 \sim \Pois{\lambda_2},\\
    (\beta \,|\,y_1, y_2) &\sim \textrm{Beta}\left(\epsilon_1 \tp y_1,\, \epsilon_2 \tp y_2\right).
\end{align}
\end{definition}

Under the Poisson-randomized representation in \cref{def:prbeta}, we
can combine \cref{eq:marglikelihood,eq:factorize} to express DNCB-MF
as~\looseness=-1
\begin{equation}
\label{eq:pmf}
    \yij{r} \tsim \textrm{Pois}\Big(\sum_{k=1}^K \thetaik{r} \phi_{kj}\Big) \; \textrm{ for } r \in \{1,2\},
\end{equation}
\begin{equation}
\label{eq:condlikelihood}
\beta_{ij} \sim \textrm{Beta}\left(\eps{1} \tp \yij{1},\, \eps{2} \tp \yij{2}\right),
\end{equation}
which connects Poisson factorization to a beta likelihood.

To complete the model, we place gamma priors over the factors, as is
standard for Poisson factorization models:\looseness=-1
\begin{align}
\label{eq:thetaprior}
    \thetaik{1},\thetaik{2} &\sim \textrm{Gam}(a_0, b_0), \\
\label{eq:phiprior}
    \phi_{kj} &\sim \textrm{Gam}(e_0, f_0).
\end{align}

\paragraph{Interpretation of the auxiliary counts.}
Intuitively, the Poisson-distributed auxiliary variables $\yij{1}$ and
$\yij{2}$ perturb the conditional distribution of $\beta_{ij}$ away
from a shared background distribution,
$\textrm{Beta}\big(\eps{1},\eps{2}\big)$. If $\yij{1} > \yij{2}$, the
distribution shifts toward values closer to 0; conversely, if $\yij{2}
> \yij{1}$, the distribution shifts toward 1. Moreover, as the overall
magnitude of $y_{ij}^{\mathsmaller{(\bullet)}} = \yij{1} \tp \yij{2}$
increases, the distribution concentrates around its mean which equals
\begin{equation}
\mathbb{E}[\beta_{ij} \,|\,\yij{1},\yij{2}] = 
\tfrac{\eps{1}+\yij{1}}{\epsilon_{0}^{\mathsmaller{(\bullet)}}+y_{ij}^{\mathsmaller{(\bullet)}}}.
\end{equation}
The effect of the non-centrality parameters (i.e., the means of the
Poisson-distributed auxiliary variables) on the DNCB marginal
distribution of $\beta_{ij}$ can be explained similarly. Because
$\mathbb{E}[\yij{r}] = \lamij{r}$ for $r \in\{1,2\}$, a large
$\lamij{1}$, relative to $\lamij{2}$, shifts the density toward 0,
while a large $\lamij{\bullet} = \lamij{1} + \lamij{2}$ concentrates
the distribution around its mean, whose functional form is in
\cref{sec:dncb_details}.

\paragraph{Interpretation of the latent factors.}
The latent factor $\phi_{kj}$ represents how relevant gene $j$ is in
latent component $k$. The largest elements of the vector
$\boldsymbol{\phi}_{k} \in \mathbb{R}_+^M$ can therefore be
interpreted as representing a ``pathway'' of genes that exhibit
correlated patterns of methylation. The latent factors $\thetaik{1}$
and $\thetaik{2}$ represent how methylated or unmethylated,
respectively, the genes in pathway $k$ are in sample $i$. As
$\thetaik{1}$ increases, relative to $\thetaik{2}$, the rate of
$\yij{1}$ increases, relative to the rate of $\yij{2}$, and the
distribution of $\beta_{ij}$ shifts toward 0. A convenient way to
jointly summarize $\thetaik{1}$ and $\thetaik{2}$ is~\looseness=-1
 \begin{equation}
 \label{eq:ratio}
 \rho_{ik} = \frac{\thetaik{1}}{\thetaik{1}+\thetaik{2}},
 \end{equation}
 where $\rho_{ik} \gg 0.5$ means pathway $k$ is hypermethylated in
 sample $i$ and $\rho_{ik} \ll 0.5$ means pathway $k$ is
 hypomethylated in sample $i$. The vector $\boldsymbol{\rho}_i \in
 (0,1)^K$ can also be interpreted as an embedding of sample $i$. We
 show these embeddings can be used to guide exploratory analyses
 in \cref{sec:predictive}.~\looseness=-1

\begin{figure*}[t!]
     \centering
     \begin{subfigure}[b]{0.32\textwidth}
         \centering
         \includegraphics[width=\textwidth]{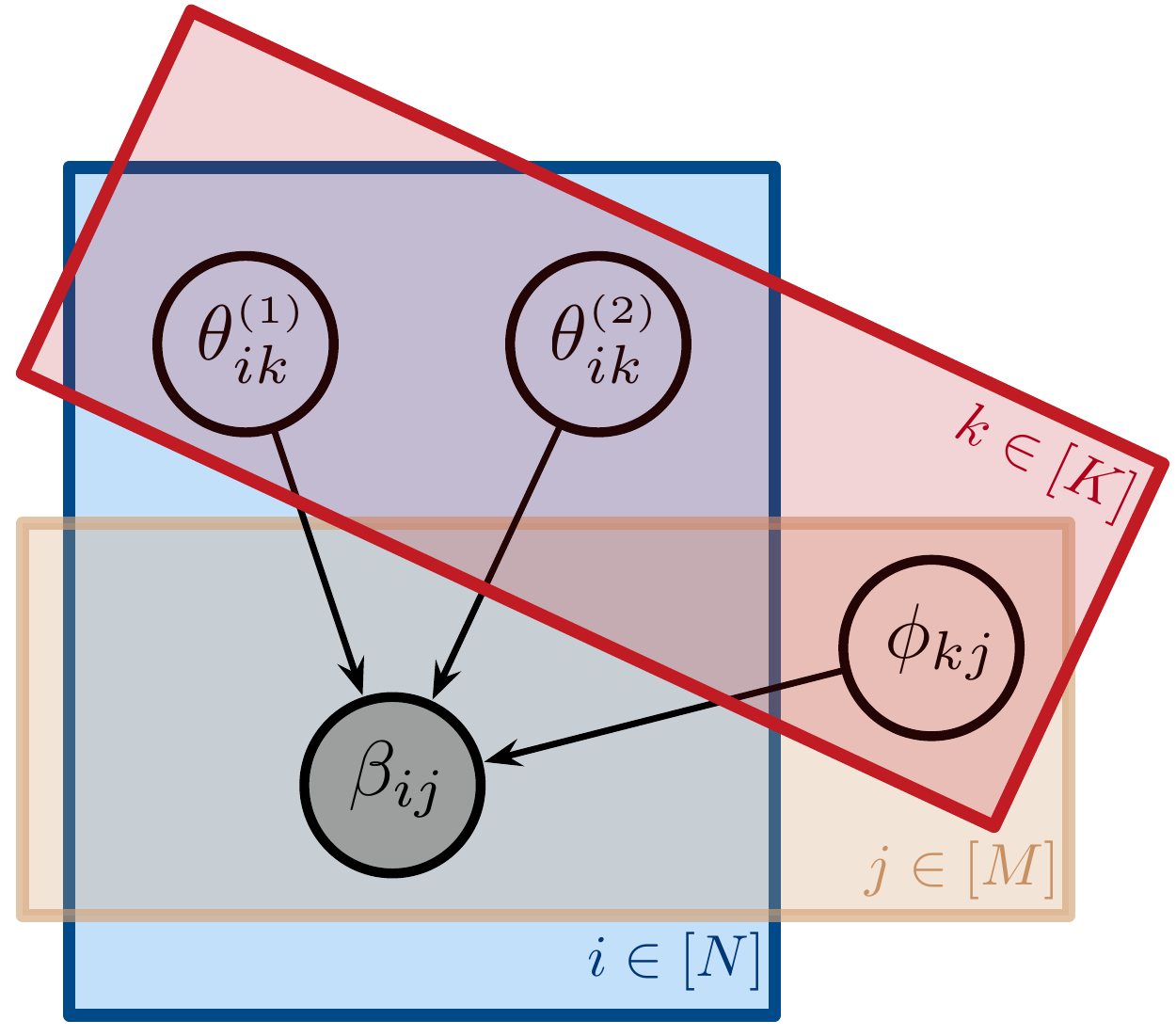}
         \caption{\footnotesize The graphical model for
           BG-NMF~\citep{ma2015variational} and for the form of DNCB-MF
           given in \cref{eq:marglikelihood}, where all of the auxiliary~variables have been marginalized out.~\looseness=-1}
         \label{fig:BG-NMF}
     \end{subfigure}
     \hfill
     \begin{subfigure}[b]{0.32\textwidth}
         \centering
         \includegraphics[width=\textwidth]{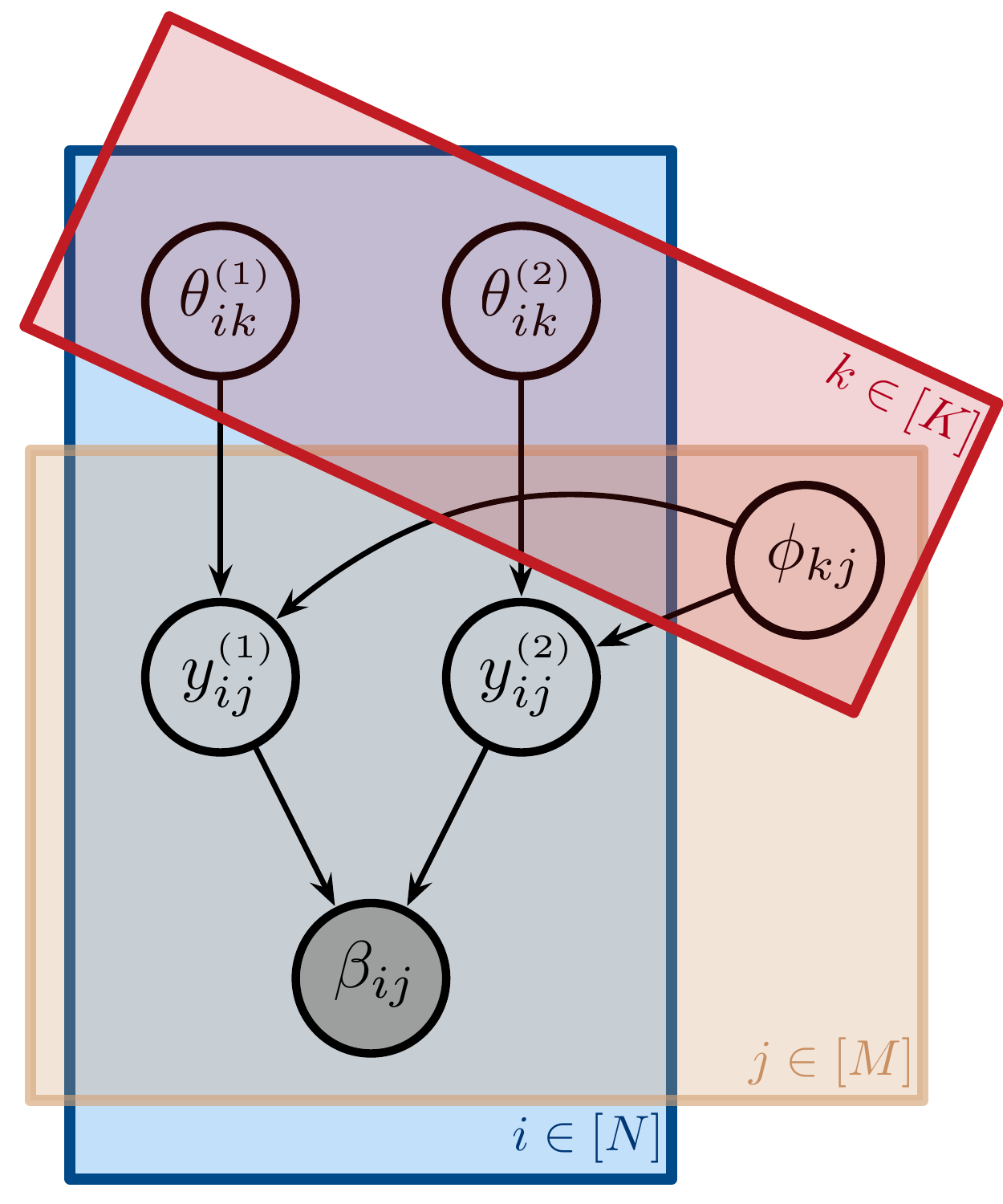}
         \caption{\footnotesize The Poisson-randomized beta form of
           DNCB-MF given in \cref{eq:pmf,eq:condlikelihood}. The
           data's dependence on the factors flows through the auxiliary
           variables $\yij{1}$ and $\yij{2}$.~\looseness=-1}
         \label{fig:dncbmf}
     \end{subfigure}
     \hfill
     \begin{subfigure}[b]{0.32\textwidth}
         \centering
         \includegraphics[width=\textwidth]{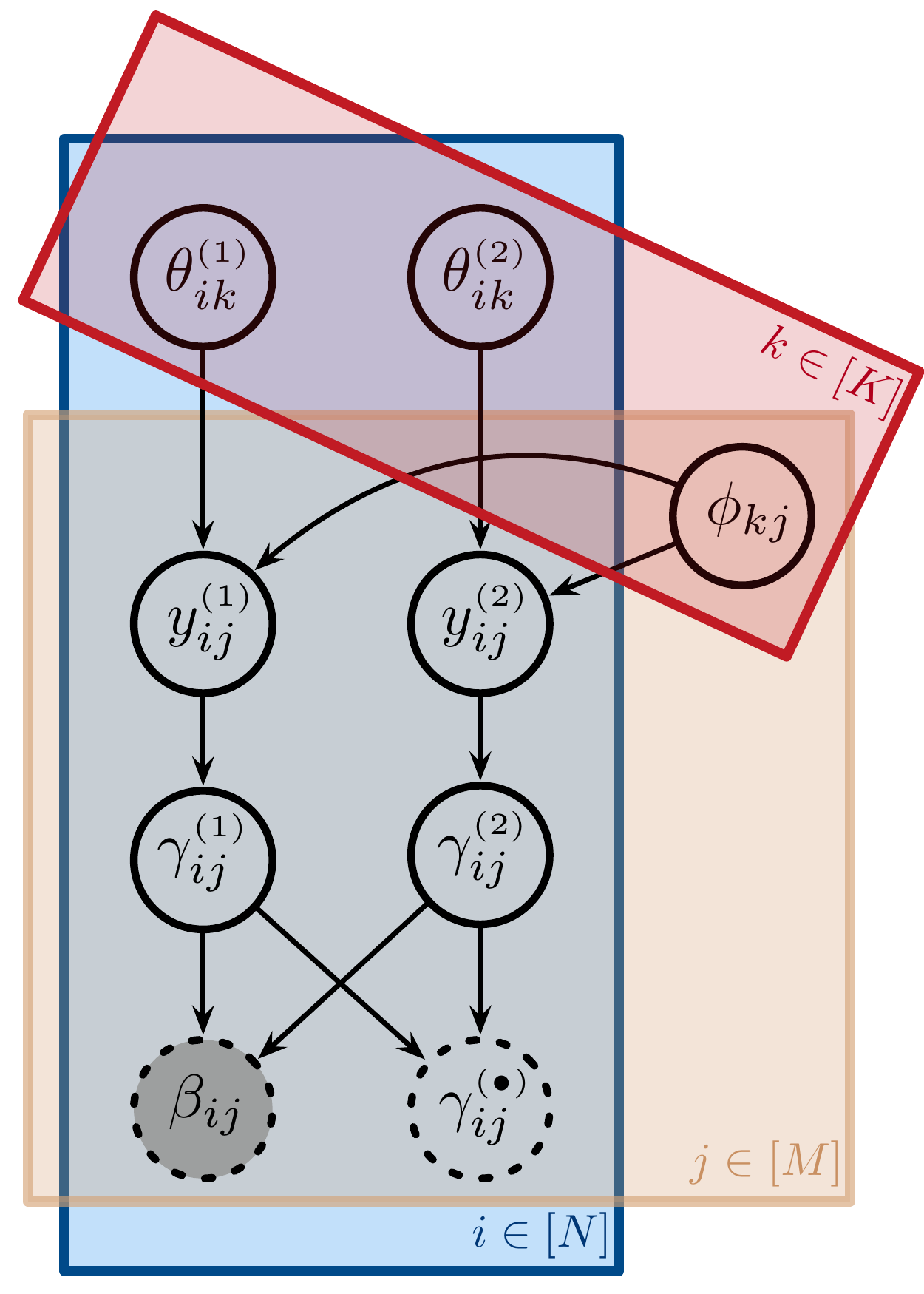}
         \caption{\footnotesize The fully augmented form of DNCB-MF
           given in \cref{eq:indepgamma,eq:sumprop}, where
           $\beta_{ij}$ is determined by $\gammaij{1}$ and
           $\gammaij{2}$. This form is particularly useful for
           posterior inference.~\looseness=-1}
         \label{fig:dncbmf-aug}
     \end{subfigure}
        \caption{\footnotesize A graphical comparison of related
          generative processes. All hyperparameters (including
          $\eps{1}$ and $\eps{2}$ for DNCB-MF) are omitted for ease of
          comparison. The plate notation represents exchangeability
          across the specified indices. Shaded nodes are observed
          variables; unshaded nodes are latent variables. Solid edges
          denote random variables; dotted edges denote deterministic
          variables.~\looseness=-1}
        \label{fig:graphical}
\end{figure*}

\section{Related work}
\label{sec:related}

In this section, we briefly review the most closely related
dimensionality reduction methods, with an emphasis on the methods that
are commonly used for DNA methylation datasets. We draw connections to
our model as appropriate.\looseness=-1

\textbf{PCA and NMF.} Non-negative matrix factorization
(NMF)~\citep{lee1999learning} factorizes an $N \times M$ matrix into
two non-negative latent factor matrices $\Theta \in \mathbbm{R}_+^{N
  \times K}$ and $\Phi \in \mathbbm{R}_+^{K \times M}$. This is
typically achieved by minimizing the Frobenius norm of the
reconstruction error subject to a non-negativity constraint on the
latent factor matrices as:\looseness=-1
\begin{equation}
    \Theta^*, \Phi^* \in \underset{\Theta, \Phi}{\textrm{argmin}}\, \lVert \mathcal{B}-\Theta \Phi \rVert_\textrm{F} \;\;\textrm{ s.t. }\;\; \Theta,\Phi \geq 0.
\end{equation}
When fit using this Frobenius loss, NMF can be viewed as performing
maximum likelihood estimation (MLE) in~a~Gaussian model that is
truncated so that $\beta_{ij} \in \mathbbm{R}_+$:\looseness=-1
\begin{equation}
  \beta_{ij} \sim \textrm{TruncNorm}(\sum_{k=1}^K \theta_{ik}\phi_{kj},\, \sigma_0).\end{equation}

Principal component analysis (PCA) involves a similar optimization
but without the non-negativity constraint on the latent factor
matrices. PCA can be viewed as performing MLE in a standard (i.e.,
non-truncated) Gaussian model.

Both PCA and NMF are commonly used in bioinformatics and have been
used for DNA methylation datasets, where NMF perform betters due to
its non-negativity
constraint~\citep{teschendorff2007elucidating,zhuang2012comparison}. In
addition, NMF is often preferred because of the ``parts-based''
interpretation of its non-negative latent factors.~\looseness=-1 

\textbf{Non-Gaussian mixture models.} Several non-Gaussian clustering
methods and mixture models have been developed specifically for DNA
methylation datasets; see~\citet{ma2014comparisons} for a survey. Some
of these models, like the recursive-partitioning beta mixture model
of~\citet{houseman2008model}, assume a beta likelihood. Although these
models make probabilistic assumptions that are appropriate for $(0,1)$
bounded-support data, they yield less expressive latent~representations
than admixture models such as NMF.~\looseness=-1

\textbf{BG-NMF.} Our model is most closely related to beta-gamma
non-negative matrix factorization (BG-NMF), which was developed by
\citet{ma2015variational} specifically for DNA methylation
datasets. BG-NMF is the first (and, to our knowledge, only) matrix
factorization model to assume a beta likelihood. Specifically, it
assumes that each element $\beta_{ij} \in (0,1)$ in a sample-by-gene
matrix is drawn as follows:\looseness=-1
\begin{align}
    \beta_{ij} &\sim \textrm{Beta}\left(\alphaij{1},\alphaij{2}\right),
\end{align}
where the two ``shape'' parameters $\alphaij{1}$ and $\alphaij{2}$ are
defined to be the same linear functions of low-rank latent factors as
those given in \cref{eq:factorize}. BG-NMF also places the same gamma
priors over these factors as those given in
\cref{eq:thetaprior,eq:phiprior}.\looseness=-1

We provide a graphical comparison of BG-NMF and DNCB-MF in
\cref{fig:graphical}. DNCB-MF and BG-NMF both factorize a sample-by-gene matrix into three non-negative latent factor
matrices; however, DNCB-MF factorizes the non-centrality parameters of
the DNCB distribution, while BG-NMF factorizes the shape parameters of the
beta distribution.\looseness=-1

Deriving an efficient and modular posterior inference algorithm for BG-NMF is
hampered by the lack of a closed-form conjugate prior for the beta
distribution. \citet{ma2015variational} propose a variational
inference algorithm that maximizes nested lower bounds on the model
evidence. Their derivation is sophisticated, but highly tailored to
the specific structure of the model, which makes the model difficult
to modify or extend. Moreover, the quality of this algorithm's
approximation to the posterior distribution is not well
understood. For biomedical settings, in which precise quantification
of uncertainty is often necessary, the lack of an efficient MCMC
algorithm therefore limits BG-NMF's applicability.\looseness=-1

\section{Posterior Inference}
\label{sec:mcmc}
\vspace{-0.55em}

Given an $N \times M$ sample-by-gene matrix of bounded-support data
$\mathcal{B} \in (0,1)^{N \times M}$, the goal is to approximate the
posterior distribution over the latent factor matrices
$P(\Theta^{\mathsmaller{(1)}}, \Theta^{\mathsmaller{(2)}}, \Phi
\,|\,\mathcal{B})$. Like the beta distribution, the DNCB distribution
lacks a closed-form conjugate prior; however, it admits several
augmentations that let us exploit special relationships between the
beta, gamma, and Poisson distributions to derive an auxiliary-variable Gibbs
sampler whose stationary distribution is the exact posterior. Moreover, this Gibbs sampler is composed entirely of closed-form complete conditionals that can be sampled from efficiently.\looseness=-1

Below, we introduce auxiliary variables that augment DNCB-MF to create
conditionally conjugate links to the latent factors. Specifically, we
work within the Poisson-randomized beta form of the model given in
\cref{eq:pmf,eq:condlikelihood}, which links $\beta_{ij}$ to a pair of
Poisson-distributed auxiliary variables $\yij{r} \sim
\textrm{Pois}(\lamij{r})$ for $r\in \{1,2\}$, whose rates $\lamij{r}$
are factorized into the latent factors. Conditioned on these auxiliary
variables, the updates for the latent factors follow from
gamma--Poisson matrix factorization~\citep{cemgil09bayesian}.

In light of this, the only thing that is needed is to derive an
efficient Gibbs sampler is a way to sample the Poisson-distributed
auxiliary variables from their complete conditionals. Our approach
relies on further augmenting the conditional likelihood using the
following definition.~\looseness=-1

\begin{definition}
A beta random variable $\beta \!\sim\! \textrm{Beta}(\alpha_1, \alpha_2)$
can be simulated as $\beta \teq
\tfrac{\gamma_1}{\gamma_1 \tp \gamma_2}$, where $\gamma_r \!\sim\!
\Gam{\alpha_r, c}$ for $r \in \{1,2\}$ are independent gamma variables
with rate $c \!>\! 0$.~\looseness=-1
\end{definition}
We can represent the conditional likelihood in
\cref{eq:condlikelihood} as\looseness=-1
\begin{align}
\label{eq:indepgamma}
    \gammaij{r} &\sim \Gam{\eps{r} \tp \yij{r}, 1}  \;\;\text{ for } r \in \{1,2\}, \\
    \label{eq:sumprop}
    &\gammaij{\bullet} = \gammaij{1}+\gammaij{2} \;\;\;\textrm{and} \;\;\; \beta_{ij} = \frac{\gammaij{1}}{\gammaij{\bullet}},
\end{align}
which corresponds to the fully augmented form of DNCB-MF shown in
\cref{fig:dncbmf-aug}. This form of our model is particularly useful
for posterior inference. Indeed, our Gibbs sampler iterates between
sampling $\yij{r}$ given $\gammaij{r}$ and vice versa.

\subsubsection*{Sampling $\gamma_{ij}^{\mathsmaller{(1)}}$ and $\gamma_{ij}^{\mathsmaller{(2)}}$}
Because the gamma-distributed auxiliary variables have a deterministic
relationship with $\beta_{ij}$, we can sample them from their complete
conditional by first sampling their sum $\gammaij{\bullet}$ from its
complete conditional and then calculating
\begin{equation}
\label{eq:compute}
    \gammaij{1} = \beta \gammaij{\bullet} \;\;\;\textrm{and}\;\;\; \gammaij{2} = (1-\beta) \gammaij{\bullet}.
\end{equation}

To derive the complete conditional of $\gammaij{\bullet}$, we appeal to the following unique property of gamma-distributed variables.~\looseness=-1
\begin{definition}
For any pair of independent positive random variables $X_1$ and $X_2$,
their sum $X_\bullet \teq X_1 + X_2$ and their proportion $\tilde{X} \teq
X_1\,/\,(X_1+X_2)$ are marginally independent---that is, $P(X_\bullet,
\tilde{X}) \teq P(X_\bullet)\,P(\tilde{X})$---if and only if $X_1$ and
$X_2$~are both gamma-distributed~\citep{lukacs1955characterization}.~\looseness=-1
\end{definition}

The complete conditional of $\gammaij{\bullet}$ is therefore \emph{independent} of $\beta_{ij}$ and equal to its distribution under the prior. Because $\gammaij{\bullet}$ is defined as the sum of two gamma-distributed random variables, its complete conditional (and prior) is as follows:
\begin{equation}
\label{eq:hallucinate}
    \gammaij{\bullet} \sim \Gam{\eps{\bullet}+\yij{\bullet}, 1}.
\end{equation}
Collectively, \cref{eq:hallucinate,eq:compute} provide an efficient
way~to~sample $\gammaij{1}$ and $\gammaij{2}$ from their complete
conditional.

\subsubsection*{Sampling $y_{ij}^{\mathsmaller{(1)}}$ and $y_{ij}^{\mathsmaller{(2)}}$}
Conditioning on $\gammaij{1}$ and $\gammaij{2}$ renders the
Poisson-distributed auxiliary variables $y_{ij}^{\mathsmaller{(1)}}$
and $y_{ij}^{\mathsmaller{(2)}}$ independent under their complete
conditional. Moreover, as shown by the following proposition, their
complete conditionals have a closed form.
\begin{definition}
If $\gamma \sim \textrm{Gam}(\epsilon \tp y, c)$ and $y \sim \Pois{\lambda}$, then the posterior of $y$ is Bessel~\citep{yuan2000bessel}:
\begin{equation}
    P(y \,|\, \gamma, \epsilon, c, \lambda) = \Bess{y; \epsilon \tm 1,\, 2\sqrt{c\gamma\lambda}},
\end{equation}
where the Bessel distribution is defined as
\begin{equation}
\Bess{y; v, a} = \frac{(\tfrac{a}{2})^{2y+v}}{y!\,\Gamma(y+v+1)\,I_v(a)}
\end{equation}
and where $I_{v}(a)$ is the first type modified Bessel function.
\end{definition}\vspace{-0.5em}
Using this definition, the complete conditional for $\yij{r}$ is
\begin{equation}
    (\yij{r}|-) \sim \Bess{\eps{r} \tm 1,\, 2\sqrt{\gammaij{r} \lamij{r}}}.
\end{equation}
\citet{devroye2002simulating} gives methods for efficiently sampling from the Bessel distribution. Although it is still relatively unknown, the Bessel distribution has gained attention in a few recent papers~\citep{zhou2015gamma,schein2019locally,schein2019poisson}.

\subsubsection*{Sampling $\thetaik{1},\thetaik{2},\phi_{kj}$}

Conditioned on $\yij{1}$ and $\yij{2}$, the updates for the latent
factors $\thetaik{1},\thetaik{2},\phi_{kj}$ follow from gamma--Poisson
matrix factorization. We provide these updates in
\cref{sec:mcmc_details}, along with a complete summary of our entire
Gibbs sampler.~\looseness=-1

\begin{figure*}[t!]
     \centering
     \includegraphics[width=\linewidth]{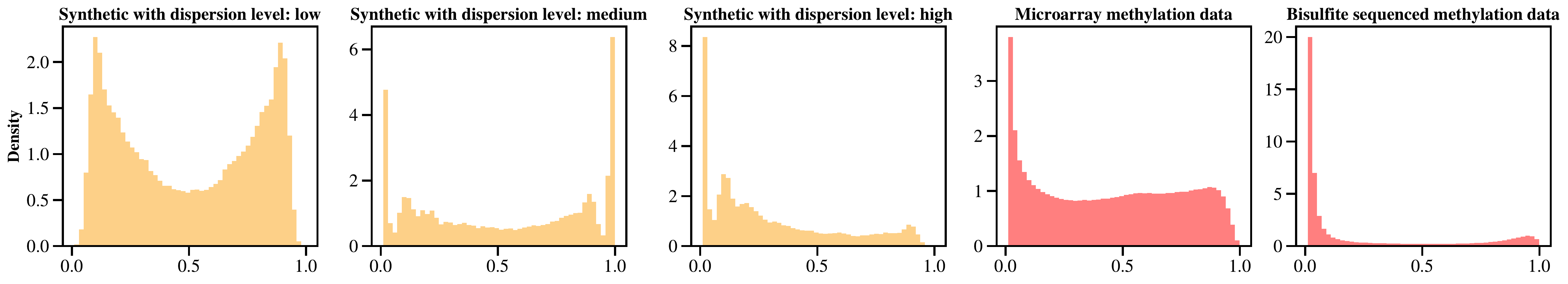}
     \caption{\footnotesize Histograms of the synthetic (yellow) and
       real (red) datasets. The synthetic datasets were created using
       Epiclomal with three levels of dispersion. As dispersion
       increases, values are pushed to the extremes. The
       high-dispersion data (middle) is most similar to the real
       data.~\looseness=-1}
     \label{fig:hist}
\end{figure*}
\begin{figure}[h!]
  \centering
  \includegraphics[width=\linewidth]{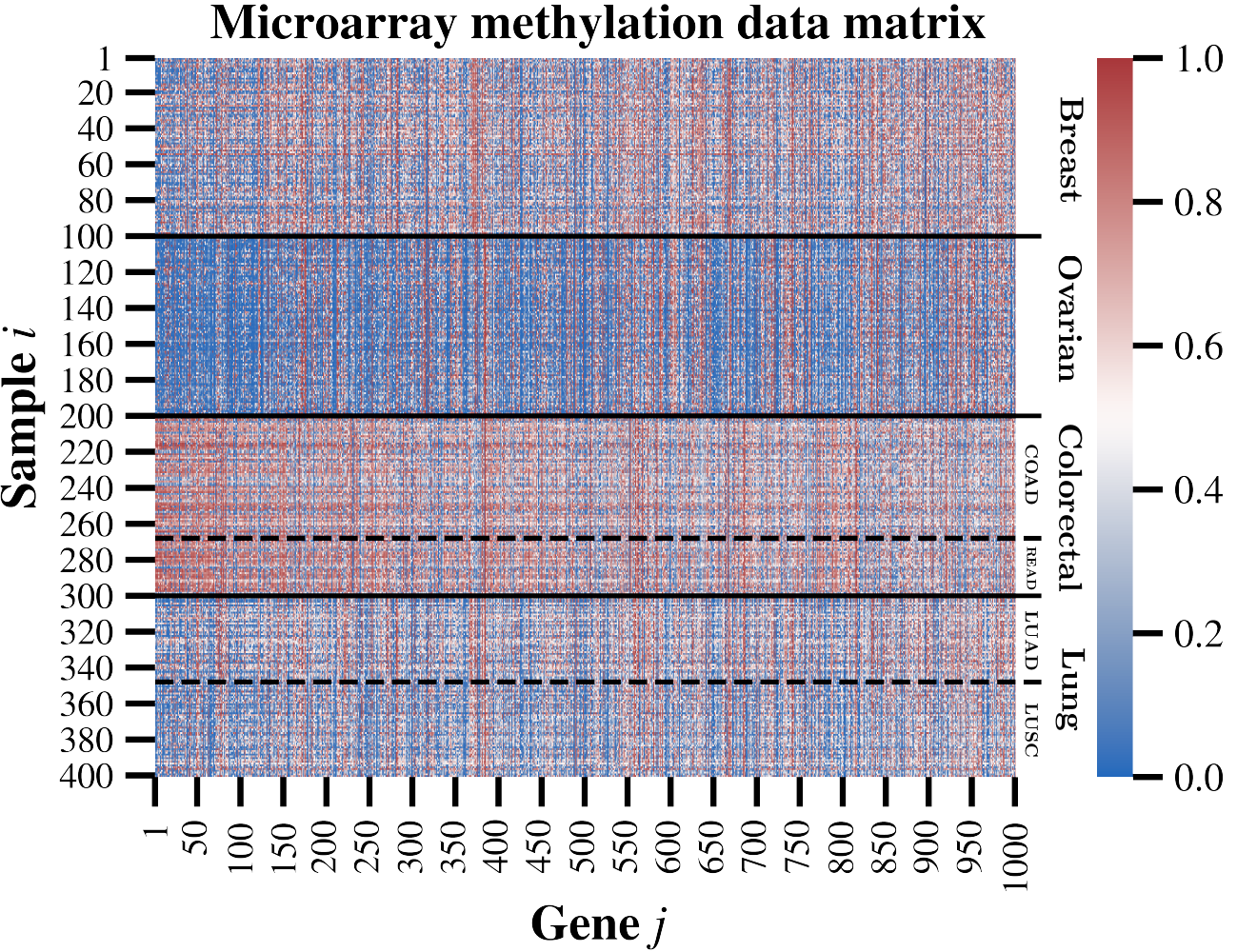}
  \caption{\footnotesize Heatmap of the microarray dataset. Only the
    first 1,000 (of $M \teq$ 5,000) genes are shown. Bright red values
    are well above 0.5 and indicate that a gene is methylated; blue
    indicates that a gene is unmethylated. The four (known) cancer
    types (100 samples from each) are annotated; they clearly display
    differentiated methylation.~\looseness=-1}
  \label{fig:dataheat}
\end{figure}

\vspace{-0.55em}
\section{Out-of-Sample Prediction}
\label{sec:predictive}
\vspace{-0.55em}

In this section, we present a study of our model's out-of-sample
predictive performance on both real and synthetic DNA methylation
datasets. We compare our model's performance to that of
state-of-the-art models in bioinformatics.

\subsection{Datasets}
\label{sec:datasets}

\paragraph{Microarray data.} We used the Cancer Genome Atlas
(TCGA)~\citep{tomczak2015cancer} to compile a dataset of 400 cancer
samples whose methylation level at about 27,000 genes was profiled
using Illumina 450K BeadChip microarrays. We selected the samples so 
that there were 100 samples each from four etiologically distinct cancer types: breast, 
ovarian, colorectal, and lung cancer. 
The colorectal and lung cancer samples further divide into two
subtypes. Although the goal of dimensionality reduction is usually to
discover novel subtypes, checking a model's ability to discover known
subtypes can be a way to assess its utility. Microarray data comes
processed into ``beta values''~\citep{kuan2010statistical}; we did not
process the data any further. Following \citet{ma2014comparisons}, we
selected the 5,000 genes with the highest variance across the samples
to obtain a $400 \times 5,000$ sample-by-gene matrix. A heatmap of
this matrix is shown in~\cref{fig:dataheat}.\looseness=-1

\paragraph{Bisulfite sequenced methylation data.} We downloaded the
dataset studied by~\citet{sheffield2017dna}. This dataset consists of
156 Ewing sarcoma cancer samples and 32 healthy samples ($N \teq
188$), whose methylation was profiled using bisulfite sequencing
(bi-seq). Bi-seq data consists of binary ``reads'' of methylation at
many loci per gene. We processed this data into ``beta values'' by
first counting all the methylated-mapped reads $d_{ij}$ and
non-methylated reads $u_{ij}$ for all loci within a given gene $j$ and
then calculating $\beta_{ij} = \frac{s_0 + d_{ij}}{2s_0 + d_{ij} +
  u_{ij}}$ with the smoothing term set to $s_0 \teq 0.1$. As with the
microarray data, we selected the 5,000 genes with the highest variance
to obtain a $188 \times 5,000$ matrix.~\looseness=-1

\textbf{Synthetic data.} To study our model's suitability for $(0,1)$
bounded-support data that may arise in other domains, we created
synthetic datasets using the Epiclomal synthetic data generator
\citep{de2020epiclomal}. Epiclomal simulates single-cell methylation
data. To create ``bulk'' data, similar to the microarray or bi-seq
data, we generated and aggregated 100 cells for every sample $i$ for
$N=100$ samples at $M=500$ genes. We varied the Epiclomal parameters
to generate datasets with three different levels of dispersion---low,
medium, and high---where increasing dispersion pushes values toward
the extremes of 0 and 1. For each level of dispersion, we generated
three datasets, all with the ``true'' number of components set to
$K^\star \teq 10$. We provide a comparison of the synthetic and real
datasets' histograms in \cref{fig:hist}.\looseness=-1

\begin{figure*}[t!]
     \centering
     \includegraphics[width=\linewidth]{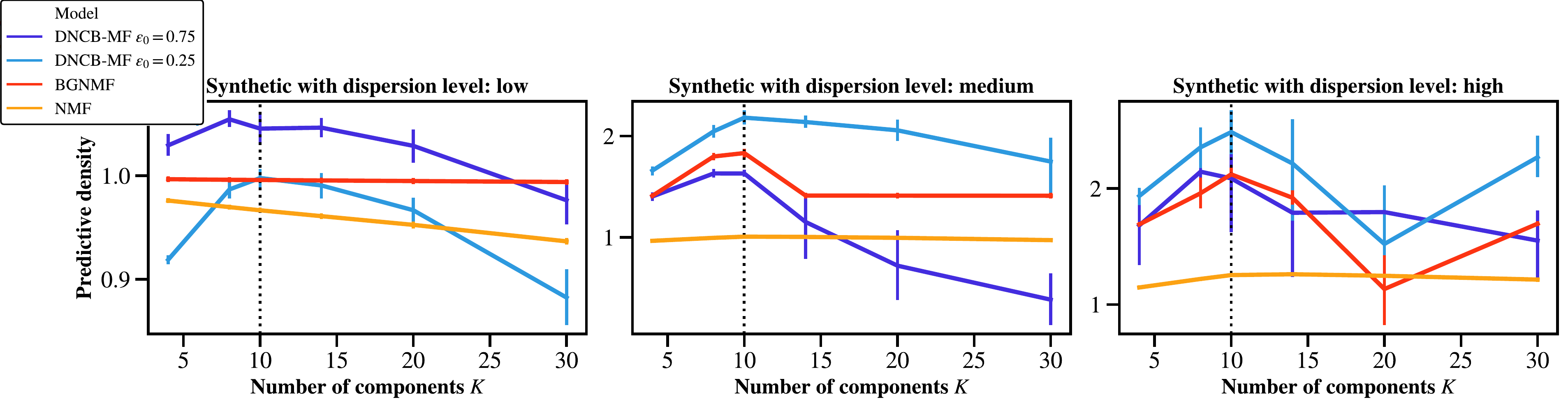}
     \caption{\footnotesize Out-of-sample predictive performance for
       the synthetic datasets. We generated datasets with three levels
       of dispersion---low (left), medium (middle), and high
       (right)---where increasing dispersion pushes values to the
       extremes of 0 and 1. For each level of dispersion, we generated
       three datasets; for each dataset, we created three train--test
       splits by generating three binary masks. For all models with
       all values of $K$, we used three different random
       initializations for each dataset and mask combination. For each
       value of $K$, we plot $\textrm{PPD}^{\frac{1}{|\mathcal{M}|}}$,
       where $|\mathcal{M}|$ is the number of held-out values,
       averaged across the initialization, dataset, and mask
       combinations; error bars indicate 95\% confidence
       intervals. For all three levels of dispersion, most models'
       performance peaks at the true number of components $K \teq
       K^\star \teq10$. NMF always performs worse than BG-NMF; BG-NMF
       is almost always ``sandwiched'' between DNCB-MF with
       $\epsilon_0 \teq 0.25$ and DNCB-MF with $\epsilon_0 \teq 0.75$
       (or vice versa). For the low-dispersion datasets, DNCB-MF with
       $\epsilon_0 \teq 0.75$ performs the best, while DNCB-MF with
       $\epsilon_0 \teq 0.25$ performs worse than than BG-NMF. For the
       medium- and high-dispersion datasets, DNCB-MF with
       $\epsilon_0=0.25$ performs the best. Intuitively, this makes
       sense: as the DNCB shape parameters get smaller, the density
       concentrates at the extremes (see
       \cref{fig:dncb}).~\looseness=-1}
     \label{fig:synthetic}
\end{figure*}
\begin{figure}[h!]
     \centering
     \includegraphics[width=0.9\linewidth]{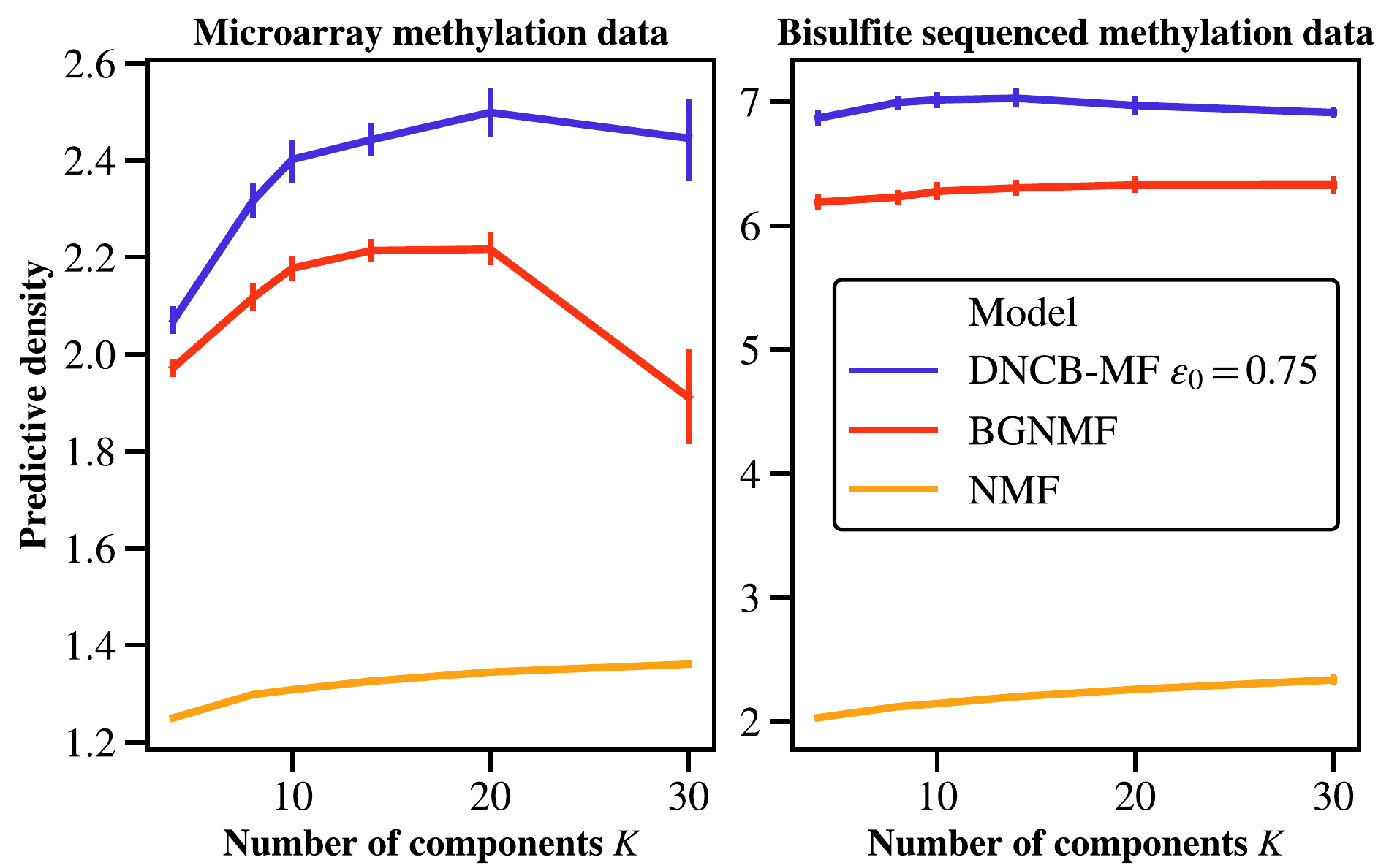}
     \caption{\footnotesize Out-of-sample predictive performance for
       the two real datasets described in \cref{sec:datasets}. For
       both, DNCB-MF with $\epsilon_0 \teq 0.75$ performs
       significantly better than NMF and BG-NMF.\looseness=-1}
     \label{fig:real}
\end{figure}
\vspace{-0.5em}

\subsection{Models}

We compare our model's out-of-sample predictive performance to that of
BG-NMF and NMF. BG-NMF is a is a state-of-the-art matrix factorization
model for DNA methylation data that differs from DNCB-MF by assuming a
beta likelihood instead of a DNCB likelihood; NMF serves as a simple
baseline because it is so commonly used.\looseness=-1

\paragraph{Setting $K$.} For all models in all
experiments, we used $K \in \{4, 8, 10, 14, 20, 30\}$. These values
are centered around $K \teq 14$, which \citet{ma2014comparisons}
report as being optimal for BG-NMF when modeling DNA methylation
datasets that~are similar in composition to our microarray dataset.~\looseness=-1

\paragraph{DNCB-MF.}
We implemented the Gibbs sampler described in \cref{sec:mcmc} in
Cython. For all experiments, we let the Gibbs sampler ``burn in'' for
1,000 iterations and then ran it for another 2,000 iterations, saving
every $20^{\textrm{th}}$ sample. We set the hyperparameters for the
gamma priors to $a_0 \teq b_0 \teq e_0 \teq f_0 \teq 0.1$ and set the
two DNCB shape parameters to $\eps{1} \teq \eps{2} \teq \epsilon_0
\teq 0.75$. For the synthetic datasets, we also experimented with
setting the shape parameters to $\epsilon_0 \teq 0.25$.~\looseness=-1

\paragraph{BG-NMF.}
We implemented BG-NMF in Python and set the hyperparameters for the
gamma priors to the values described above for DNCB-MF. We set all
other hyperparameters to the values recommended
by~\citet{ma2014comparisons}.

\paragraph{NMF.}
We used the implementation of NMF in
Scikit-learn~\citep{pedregosa2011scikit} with the default settings.

\paragraph{Code.}
We have released our implementations of DNCB-MF and
BG-NMF.\footnote{\url{https://github.com/aschein/dncb-mf}} Our code includes fast samplers for
the Bessel distribution and an algorithm for computing the density of
the DNCB distribution. We have also released the real and synthetic
datasets that we used for our experiments.

\subsection{Study design}

\paragraph{Random masks.}
We created three train--test splits for each dataset (real or
synthetic) by generating three binary masks $\mathcal{M}$ that ``hold
out'' a random 10\% of the sample-by-gene matrix. For all models with
all values of $K$, we used three different random initializations for
each dataset and mask combination. All models took the mask as input
and imputed the held-out values during inference.\looseness=-1

\paragraph{Evaluation metric.} To assess out-of-sample predictive performance, we used the pointwise predictive density (PPD)~\citep{gelman2014understanding}. For NMF, this is as follows:\looseness=-1
\vspace{-0.1em}
\begin{equation}
    \textrm{PPD}_{\textrm{point}} = \prod_{i,j \in \mathcal{M}}
    P\big(\beta_{ij} \,|\, \Theta^*, \Phi^*\big),
\end{equation}
where $\Theta^*$ and $\Phi^*$ are point estimates of NMF's latent
factor matrices and $P(\cdot\,|\,\cdot)$ denotes a Gaussian truncated
to $(0,\infty)$.~\looseness=-1

For BG-NMF and DNCB-MF, the PPD is given by
\begin{equation}
\label{eq:ppd}
    \textrm{PPD}_{\textrm{post}} = \prod_{i,j \in \mathcal{M}} \Big[\tfrac{1}{S} \sum_{s=1}^S P\big(\beta_{ij} \,|\, \Theta^{(1)}_s, \Theta^{(2)}_s, \Phi_s\big)\Big],
\end{equation}
where $\Theta^{(1)}_s, \Theta^{(2)}_s, \Phi_s$ are samples from the
posterior distribution, either saved during MCMC for DNCB-MF or drawn
from the fitted variational distribution for BG-NMF. For both models,
we used $S\teq 100$. The predictive density $P(\cdot|\cdot)$ is the
beta distribution for BG-NMF and the DNCB distribution for our
model. Computing the DNCB density requires computing Humbert's
confluent hypergeometric function, for which we implemented the
algorithm of~\citet{orsi2017new}.

For all models, we ultimately report
$\textrm{PPD}^{\frac{1}{|\mathcal{M}|}}$, where $|\mathcal{M}|$ is the
number of held-out values, which is equivalent to the geometric mean
of the predictive densities across the held-out values and is
therefore comparable across all experiments. We note that this scaled version of PPD is the inverse of perplexity, an evaluation metric that is commonly used to~evaluate statistical topic models and language models.~\looseness=-1

\vspace{-0.25em}
\subsection{Results}
\vspace{-0.25em}
Out-of-sample predictive performance for the synthetic and real
datasets are shown in \cref{fig:synthetic,fig:real}, respectively. Our
model performs significantly better than NMF and BG-NMF. The results
for the synthetic datasets reveal an intuitive relationship between
the level of dispersion and the two DNCB shape parameters
$\epsilon_0^{(1)} = \epsilon_0^{(2)} = \epsilon_0$, with a smaller
parameter value yielding better performance for the more highly
dispersed datasets. For these datasets, DNCB-MF with $\epsilon_0 =
0.25$ improves performance over both NMF and BG-NMF.\looseness=-1

\begin{figure*}
     \centering
     \begin{subfigure}[b]{0.46\textwidth}
        \includegraphics[width=\linewidth]{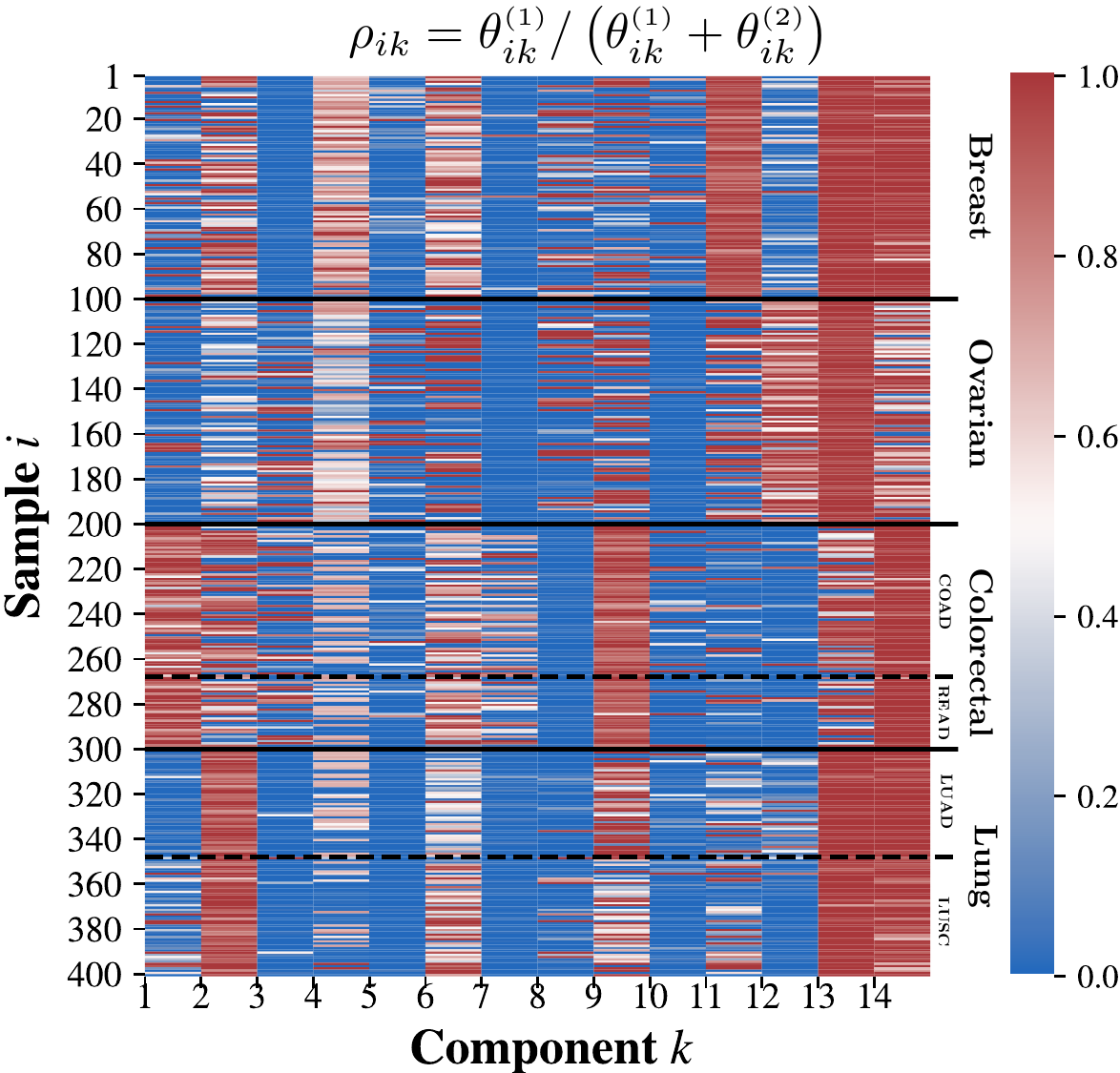}
        \caption{\footnotesize Heatmap of the embedding matrix, as defined in \cref{eq:ratio}.}
        \label{fig:array_embedding}
     \end{subfigure}
     \hfill
     \begin{subfigure}[b]{0.46\textwidth}
         \centering
         \includegraphics[width=\linewidth]{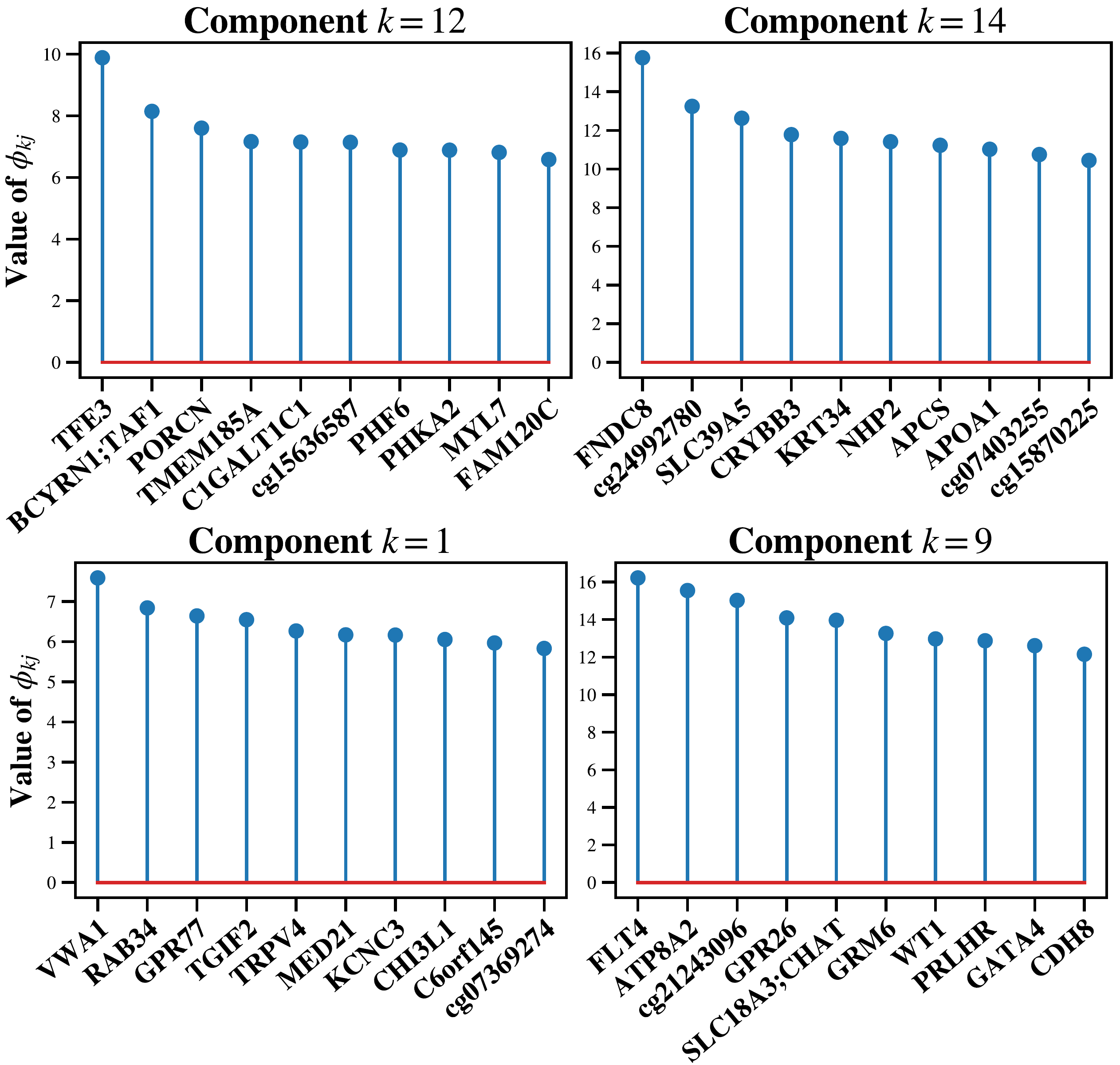}
         \caption{\footnotesize The top ten genes for four components.}
         \label{fig:components}
     \end{subfigure}
     \hfill
        \caption{\footnotesize The latent representations discovered by DNCB-MF when applied to the microarray dataset. \emph{Left:} The inferred embedding matrix. The dataset contains four cancer types (separated by horizontal lines); two of these cancer types (colorectal and lung) further divide into two subtypes (separated by horizontal lines). \emph{Right:} The genes with the ten highest $\phi_{kj}$ values for four components $k = 1,9,12,14$.}\vspace{-1em}
        \label{fig:structure}
\end{figure*}

\vspace{-0.55em}
\section{Case study}
\label{sec:qualitative}
\vspace{-0.55em}

Here, we explore our model's ability to discover meaningful latent
representations by applying our model to the microarray dataset
described in \cref{sec:datasets} and exploring the resulting
representations. The microarray dataset contains samples from four
different cancer types: breast, ovarian, colorectal, and lung cancer;
the colorectal and lung cancer samples further divide into two
subtypes (see~\cref{fig:dataheat}). We find that the latent
representations discovered by our model accord with existing
epigenetic knowledge about the gene pathways that play major roles in
the six different cancer types.\looseness=-1

\Cref{fig:array_embedding} contains the inferred embedding matrix,
where each row $\boldsymbol{\rho}_i$ is an embedding of sample $i$, as
defined in \cref{eq:ratio}. A red value denotes $\rho_{ik} > 0.5$ and
indicates that the loci proximal to the genes relevant to component
$k$ are hypermethylated in sample $i$ according to the model;
conversely, a blue value indicates that the loci are hypomethylated
according to the model. In \cref{fig:components}, we additionally show
the genes with the ten highest $\phi_{kj}$ values for four
components.~\looseness=-1

Component $k \teq 1$ has red (high) values for the colorectal cancer
samples and blue (low) values for the lung cancer samples, suggesting
that the top genes for that component are hypermethylated in
colorectal cancer and hypomethylated in lung cancer, respectively. The
component stem plot in \cref{fig:components} shows that the top genes
include \emph{RAB34}, a member of the Ras oncogene
family~\citep{sun2018}. In general, DNA methylation leads to gene
silencing (especially when proximal to the promoter region) and
therefore hypomethylation can activate oncogenes like
\emph{RAB34}~\citep{moore2013}.~\looseness=-1

The colorectal cancer samples exhibit hypermethylation for component
$k \teq 9$, whose top gene is \emph{FLT4}. Recent work demonstrates
that suppressing \emph{FLT4} inhibits cancer
metastasis~\citep{xiao2015}; it is a known therapeutic
target.\looseness=-1

All samples exhibit hypomethylation for component $k \teq 12$, except
for the ovarian cancer samples. The top gene is \emph{TFE3}, which
promotes activation of the transforming growth factor beta
(TGF$\beta$) signaling pathway. \emph{TFE3} translocation and
subsequent activation is a well-known cause of adult renal cell
carcinoma~\citep{sukov2012}.\looseness=-1

Conversely, all samples exhibit hypomethylation for component $k \teq
12$, except for the ovarian cancer samples. The top gene is the
fibronectin protein \emph{FNDC8}. Fibronectin promotes cell migration
and invasion in ovarian cancer~\citep{yousif2014}, so this accords
with existing biological knowledge.\looseness=-1

\vspace{-0.55em}
\section{Conclusion}
\vspace{-0.55em}

We presented DNCB-MF, a new non-negative factorization model for
$(0,1)$ bounded-support data based on the DNCB distribution. The DNCB
distribution is an attractive alternative to the beta distribution. As
well as being more expressive, the DNCB distribution admits several
augmentations that connect DNCB-MF to Poisson factorization models,
which are well studied and easy to build on. Although DNCB-MF was
developed specifically for DNA methylation data, the model structure
is sufficiently general that it can be adapted to other domains.  We
showed that DNCB-MF improves out-of-sample predictive performance on
both real and synthetic DNA methylation datasets over state-of-the-art
methods in bioinformatics and that the resulting representations
accord with existing epigenetic knowledge.\looseness=-1

\section{Acknowledgements}
This work was supported in part by National Science Foundation Award 1934846. We also thank Mingyuan Zhou and Daniel Sheldon for their work on a precursor to this model.

\balance
\bibliography{schein_713}
\pagebreak
\clearpage

\appendix
\nobalance
\section{The DNCB distribution}
\label{sec:dncb_details}
\begin{figure}[t!]
  \centering
  \includegraphics[width=\linewidth]{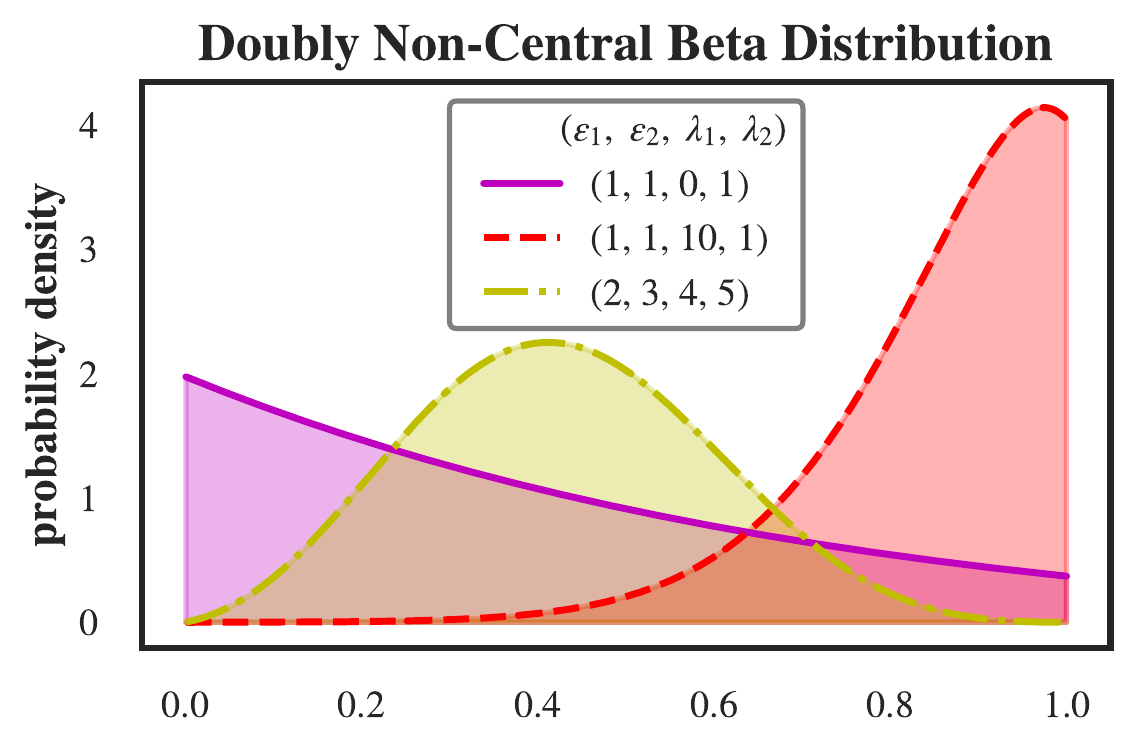}
  \caption{\footnotesize With $\epsilon_1, \epsilon_2 \geq 1$, the DNCB distribution takes unimodal shapes over $(0,1)$, similar to the beta distribution; with $\lambda_1=\lambda_2=0$, the DNCB distribution coincides with the beta distribution.}\label{fig:dncb_unimodal}
\end{figure}

The DNCB distribution is defined in \cref{def:dncb}. It can take the
same set of unimodal shapes over the $(0,1)$ interval as the beta
distribution (see \cref{fig:dncb_unimodal}), as well as multi-modal
shapes when the shape parameters $\epsilon_1, \epsilon_2 < 1$ (see
\cref{fig:dncb}).\looseness=-1

\citet{ongaro2015some} provide a general formula for the moments of
the DNCB distribution. Its first moment is
\begin{align*}
  &\mathbb{E}[\beta] =\\
  &\quad \tfrac{\epsilon_1}{\epsilon_\bullet e^{\lambda_\bullet}}  \left[{}_1F_1\left(\epsilon_\bullet; \epsilon_\bullet \tp 1; \lambda_\bullet \right) + \tfrac{\epsilon_\bullet \lambda_1}{\epsilon_1(\epsilon_\bullet \tp 1)}{}_1F_1\left(\epsilon_\bullet \tp 1; \epsilon_\bullet \tp 2; \lambda_\bullet \right)\right]
\end{align*}
where ${}_1F_1(\cdot; \cdot; \cdot)$ denotes Kummer's confluent
hypergeometric function. The second moment is more involved, but also does
not involve any special functions beyond ${}_1F_1(\cdot; \cdot;
\cdot)$.

Computing the mean and variance of the DNCB is easy because there are
many efficient open-source implementations of ${}_1F_1(\cdot; \cdot;
\cdot)$---e.g., in the Python library
scipy~\citep{virtanen2020scipy}. On the other hand, computing the DNCB
density, which we need to assess out-of-sample predictive performance
(see~\cref{sec:predictive}), requires computing Humbert's confluent
hypergeometric function $\Psi_2[\cdot; \cdot, \cdot; \cdot, \cdot]$
for which we know of no open-source implementations. We therefore
implemented the algorithm of \citet{orsi2017new} in Cython. We have
released our code for this, along with our implementations of DNCB-MF
and BG-NMF and the real and synthetic datasets that we used for our
experiments.\footnote{\url{https://github.com/aschein/dncb-mf}}~\looseness=-1
\vfill

\section{Posterior Inference}
\label{sec:mcmc_details}

Here, we provide a complete summary of our entire Gibbs sampler. As we
described in \cref{sec:mcmc}, the first step is to sample the
gamma-distributed auxiliary variables:
\begin{align}
    \label{eq:firststep}
    \compcond{\gammaij{\bullet}} &\sim \Gam{\eps{\bullet}+\yij{\bullet}, 1},\\
    \gammaij{1} = \beta \gammaij{\bullet} &\;\;\;\textrm{and}\;\;\; \gammaij{2} = (1-\beta) \gammaij{\bullet}.
\end{align}

The Poisson-distributed auxiliary variables are then conditionally
independent Bessel random variables---i.e.,
\begin{equation}
    (\yij{r}|-) \sim \Bess{\eps{r} \tm 1,\, 2\sqrt{\gammaij{r} \sum_{k=1}^K \thetaik{r}\phi_{kj}}}
\end{equation}
for $r \in \{1,2\}$. Conditioned on these auxiliary counts, the
updates for the latent factors follow from gamma--Poisson matrix
factorization. First, we represent each count as the sum of $K$
subcounts---i.e.,
$\yij{r}=\sum_{k=1}^Ky_{ijk}^{\mathsmaller{(r)}}$. By Poisson
additivity, each of these subcounts is Poisson distributed and their
complete conditional is a multinomial distribution:\looseness=-1
\begin{equation}
    \compcond{\big(y_{ijk}^{\mathsmaller{(r)}}\big)_{k=1}^K} \sim \Multi{\yij{r},\, \left(\tfrac{\thetaik{r}\phi_{kj}}{\sum_{k'=1}^K \theta_{ik'}^{\mathsmaller{(r)}}\phi_{kj}}\right)_{k=1}^K}.
\end{equation}
By Poisson--gamma conjugacy, the complete conditionals of the latent
factors, conditioned on the subcounts, are
\begin{align}
    &\compcond{\thetaik{r}} \sim \Gam{a_0 \tp \sum_{j=1}^M y_{ijk}^{\mathsmaller{(r)}},\, b_0 \tp \sum_{j=1}^M \phi_{kj}},\\
    \label{eq:laststep}
    &\compcond{\phi_{kj}} \sim\notag\\
    &\quad \Gam{e_0 \tp \sum_{i=1}^N \sum_{r=1}^2 y_{ijk}^{\mathsmaller{(r)}},\, f_0 \tp \sum_{i=1}^N \sum_{r=1}^2 \thetaik{r}}.
\end{align}

\Crefrange{eq:firststep}{eq:laststep} summarize the entire Gibbs
sampler for DNCB-MF. Iteratively following these steps is
asymptotically guaranteed to sample from the exact
posterior.\looseness=-1

\end{document}